\documentclass[sigconf]{acmart}
\pdfoutput=1
\AtBeginDocument{%
  \providecommand\BibTeX{{%
    \normalfont B\kern-0.5em{\scshape i\kern-0.25em b}\kern-0.8em\TeX}}}

\copyrightyear{2022} 
\acmYear{2022} 
\setcopyright{acmcopyright}\acmConference[CIKM '22]{Proceedings of the 31st ACM International Conference on Information and Knowledge Management}{October 17--21, 2022}{Atlanta, GA, USA}
\acmBooktitle{Proceedings of the 31st ACM International Conference on Information and Knowledge Management (CIKM '22), October 17--21, 2022, Atlanta, GA, USA}
\acmPrice{15.00}
\acmDOI{10.1145/3511808.3557718}
\acmISBN{978-1-4503-9236-5/22/10}

\usepackage{tikz}
\usepackage{balance}
\usepackage{pgfplots}
\pgfplotsset{compat=1.17}

\begin{document}

\title{Unified Knowledge Prompt Pre-training for Customer Service Dialogues}


\author{Keqing He}
\affiliation{%
  \institution{Meituan Group}
  \city{Beijing}
  \country{China}}
\email{kqin@bupt.cn}

\author{Jingang	Wang}
\affiliation{%
  \institution{Meituan Group}
  \city{Beijing}
  \country{China}}
\email{wangjingang@meituan.com}

\author{Chaobo Sun}
\affiliation{%
  \institution{Meituan Group}
  \city{Beijing}
  \country{China}}
\email{sunchaobo@meituan.com}

\author{Wei Wu}
\affiliation{%
  \institution{Meituan Group}
  \city{Beijing}
  \country{China}}
\email{wuwei1988@gmail.com}







\renewcommand{\shortauthors}{Keqing He, Jingang Wang, Chaobo Sun, \& Wei Wu}

\begin{abstract}
Dialogue bots have been widely applied in customer service scenarios to provide timely and user-friendly experience. These bots must classify the appropriate domain of a dialogue, understand the intent of users, and generate proper responses. Existing dialogue pre-training models are designed only for several dialogue tasks and ignore weakly-supervised expert knowledge in customer service dialogues. In this paper, we propose a novel unified knowledge prompt pre-training framework, UFA (\textbf{U}nified Model \textbf{F}or \textbf{A}ll Tasks), for customer service dialogues. We formulate all the tasks of customer service dialogues as a unified text-to-text generation task and introduce a knowledge-driven prompt strategy to jointly learn from a mixture of distinct dialogue tasks. We pre-train UFA on a large-scale Chinese customer service corpus collected from practical scenarios and get significant improvements on both natural language understanding (NLU) and natural language generation (NLG) benchmarks.
\end{abstract}




\maketitle

\section{Introduction}
Dialogue bots have been widely applied in most customer service scenarios like Amazon and Meituan to reduce labor cost and improve user experience. These chatbots interact with customers to introduce products, answer questions and negotiate, etc. A practical chatbot has to 
classify the appropriate domain of a dialogue, understand the intent of users, and generate proper responses. Unlike regular language understanding and generation of plain text, dialogues are often informal, back channeling, reconfirmations, hesitations, and speaker interruptions, which makes it challenging to directly apply existing pre-trained language models (PLM) \cite{devlin-etal-2019-bert,Radford2018ImprovingLU,Liu2019RoBERTaAR,NEURIPS2019_dc6a7e65,DBLP:conf/iclr/LanCGGSS20} to dialogues. It's valuable to explore specific pre-trained LM for dialogues, especially customer service dialogues.

Existing research efforts on dialogue PLM are classified into two types, understanding and generation. The former equip BERT \cite{devlin-etal-2019-bert} with dialogue-specific pre-training tasks like utterance order prediction \cite{Gu2021DialogBERTDR,Zhang2021StructuralPF}, masked speaker prediction \cite{Wang2021CSBERTAP}, response selection \cite{Wu2020TODBERTPN,henderson-etal-2020-convert}, disturbance identification \cite{Zhang2021DialogueBERTAS}. These methods focus on language understanding tasks, such as domain classification and intent detection, which limits broader application to generative tasks. In contrast, another line of researches \cite{Zhang2020DIALOGPTL,Zhong2021DialogLMPM} generates dialogue responses by further pre-training GPT \cite{Radford2018ImprovingLU} or UNILM \cite{Dong2019UnifiedLM}. Although these dialogue PLMs can generate plausible responses, it's hard to control and convey knowledgeable outputs. Typical customer service scenarios require high-quality and reasonable responses, so a dialogue PLM must understand the background knowledge of dialogue like domain and user intent. Generally, existing dialogue PLMs only aim at parts of downstream dialogue tasks, hardly improving the overall performance of a dialogue system. Besides, they focus on constructing diverse unsupervised pre-training tasks but ignore weakly-supervised dialogue knowledge in practical customer service dialogues
. Unifying all the dialogue tasks by integrating dialogue knowledge is far from well-explored.

To address the above issues, in this paper, we propose a novel unified knowledge prompt pre-training framework, UFA (\textbf{U}nified Model \textbf{F}or \textbf{A}ll Tasks), for customer service dialogues. Our method formulates all the tasks of customer service dialogues as a unified text-to-text generation task, including domain classification, intent detection, dialogue generation and summarization. This architecture simplifies the difficulty of developing and deploying different models. To utilize dialogue knowledge, we propose a knowledge-driven prompt strategy to combine distinct dialogue tasks, which helps disentangle task dependency from each other. Orthogonal to existing work, UFA aims to learn expert knowledge related to business rules from weakly-annotated dialogue labels instead of linguistic knowledge from unsupervised dialogue corpus.\footnote{In this paper, we take both the learning mechanisms from unsupervised or supervised data as pre-training to distinguish downstream finetuning.} We train our UFA with over 750 million Chinese dialogues collected from practical conversations between users and customer service agents on an E-commerce platform, Meituan. To verify the effectiveness of UFA, we conduct extensive experiments on downstream dialogue benchmarks, including natural language understanding (NLU) and natural language generation (NLG) tasks. Results show UFA significantly outperforms the baselines by 2.69\% accuracy in NLU tasks and 4.79\% Rouge-1 score in NLG tasks. Further few-shot learning analysis proves UFA has stronger capability under low-resource settings. We also perform ablation studies to demonstrate the effectiveness of each pre-training task and prompt strategy. 


Our contributions are three-fold:
\begin{itemize}
    \item To the best of our knowledge, we are the first to propose a unified pre-training framework combining all dialogue tasks for customer service dialogues. The unified architecture simplifies the difficulty of practical deployment.
    \item We propose a knowledge-driven prompt strategy to combine distinct dialogue tasks. The prompt helps bridge connections between different tasks and disentangle task dependency via task descriptions.
    \item We pre-train UFA on a large-scale customer service corpus collected from practical scenarios and perform extensive experiments on downstream benchmarks. The results show the effectiveness of our proposed model, especially in the few-shot learning.
\end{itemize}

\section{Methodology}
\subsection{Pre-training Corpus}
We collect human-human conversations where a user interacts with a human agent. Apart from these raw dialogues, we also obtain corresponding weakly-annotated labels such as domain, intent information, and summary. Note that these annotated labels are mostly generated by existing neural models like TextCNN \cite{kim-2014-convolutional} or Transformer \cite{vaswani2017attention} and then partly renewed by human agents. We don't consider human-machine conversations because the data quality is poor. Finally, we get 750 million Chinese dialogues along with related expert labels for one year from all businesses on a large E-commerce platform, Meituan. 

\begin{figure}[t]
    \centering
    \resizebox{.45\textwidth}{!}{
    \includegraphics{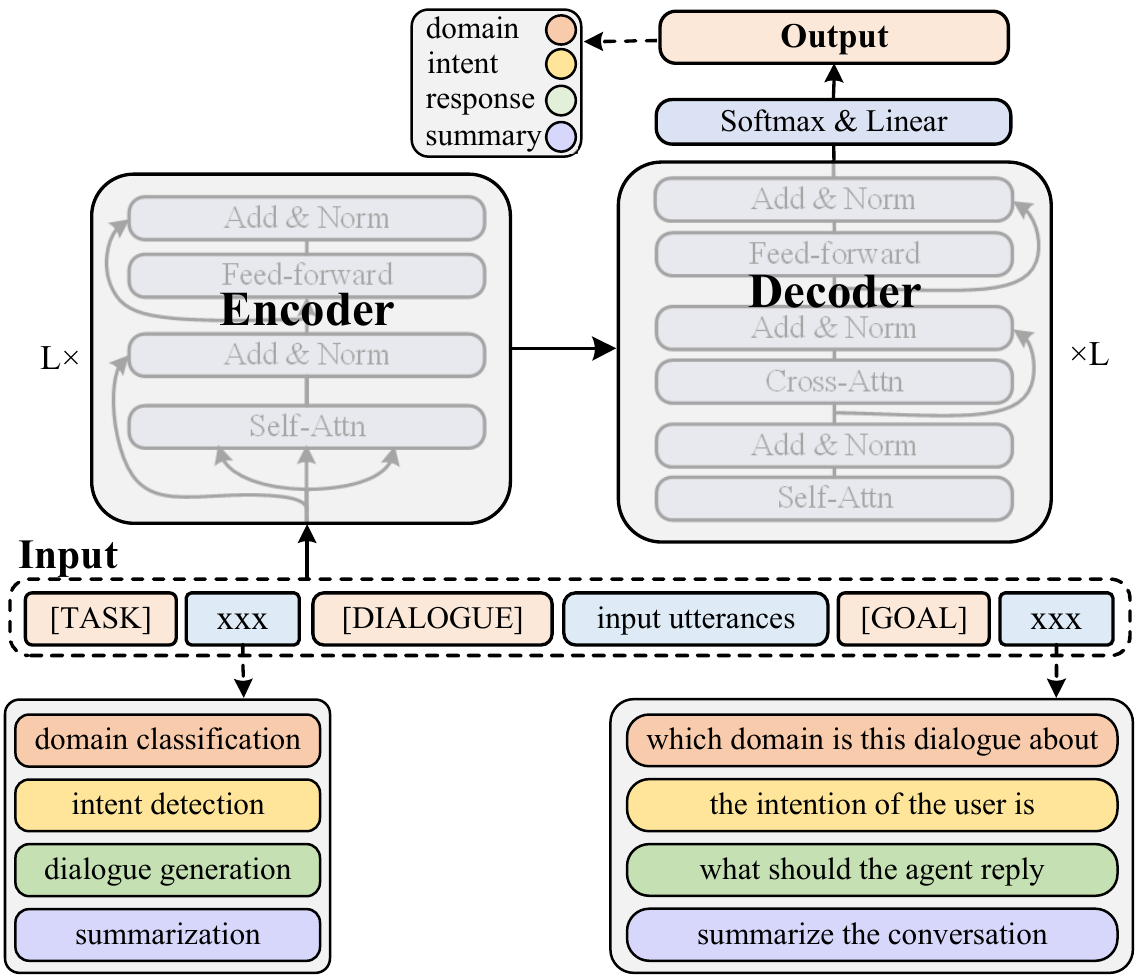}}
    \vspace{-0.35cm}
    \caption{The overall architecture of our proposed unified knowledge prompt pre-training framework (UFA) for customer service dialogues. We formulate all the dialogue tasks as a text-to-text format and use the same prompts to guide multi-task pre-training on a large-scale weakly-supervised customer service dialogue corpus.}
    \label{model}
     \vspace{-0.45cm}
\end{figure}

\subsection{Model Architecture}
Different from existing work \cite{Zhang2021DialogueBERTAS,Wang2021CSBERTAP} based on BERT \cite{devlin-etal-2019-bert}, we employ a Transformer Encoder-Decoder architecture \cite{vaswani2017attention} following \citet{Raffel2020ExploringTL}, as shown in Fig \ref{model}. Our intuition is to convert all dialogue tasks into a text-to-text format so we can use a unified model to accomplish the whole customer service dialogue system. We firstly pre-train our model using span denoising loss \cite{Raffel2020ExploringTL} on unlabeled dialogue corpus to warm up model parameters. Then we perform UFA pre-training using these weakly-annotated dialogue labels. Specifically, we concatenate dialogue history and individual task prompts which we will describe later as input, and predict corresponding task labels. We employ 8-layer Transformer blocks \cite{vaswani2017attention} both for the encoder and decoder, and each block has 6 attention heads. The output of the final decoder block is fed into a dense layer with a softmax output. We iteratively optimize all dialogue tasks including domain classification, intent detection \cite{He2020SyntacticGC, Xu2020ADG}, dialogue generation \cite{Zhang2020DIALOGPTL} and summarization \cite{Lei2021HierarchicalSS,Lei2021AFU,Zhao2022DomainOrientedPT} in an autoregressive way.

\begin{table*}[t]
\caption{Performance comparison of different models on
four dialogue tasks. UFA-ori denotes we pre-train UFA model using span denoising loss \cite{Raffel2020ExploringTL} on unsupervised dialogue corpus. UFA represents we perform unified knowledge prompt pre-training using weakly-annotated labels. For fair comparison, we use baseline models with similar parameters as our UFA. We only report metrics of bert-based models on NLU tasks. The bold numbers means statistically significant difference with $p < 0.01$.}
\vspace{-0.3cm}
\label{main}
\resizebox{0.90\textwidth}{!}{
\begin{tabular}{l|c|c|c|cc|ccc}
\hline
     Models   & Parameters & \multicolumn{1}{c|}{Domain Classification} & \multicolumn{1}{c|}{Intent Detection} & \multicolumn{2}{c|}{Dialogue Generation}    & \multicolumn{3}{c}{Summarization}                \\ 
              &            & Acc                        & Acc                        & Bleu-2         & Rouge-1        & Rouge-1        & Rouge-2        & Rouge-l        \\ \hline
bert-base     & 110M       & 89.16                      & 82.01                      & -              & -              & -              & -              & -              \\
roberta-large & 330M       & 89.12                      & 81.21                      & -              & -              & -              & -              & -              \\
mT5     & 300M       & 87.31                      & 76.75                      & 20.55          & 41.33          & 24.41          & 17.96          & 24.65          \\ \hline
UFA-ori       & 220M       & 88.46                      & 79.44                      & 27.20          & 45.93          & 25.44          & 18.58          & 25.05          \\ 
UFA           & 220M       & \textbf{91.09}             & \textbf{85.45}             & \textbf{29.66} & \textbf{48.30} & \textbf{27.02} & \textbf{20.10} & \textbf{26.63} \\ \hline
\end{tabular}}
\vspace{-0.25cm}
\end{table*}

\subsection{Model Input and Output Format}
In order to train a unified model on the diverse set of dialogue tasks described above, we cast all of the tasks we consider into a “text-to-text” format—that is, a task where the model is fed some text for context or conditioning and is then asked to predict some output text. This framework provides a consistent training objective both for pre-training and fine-tuning. To identify which task the model should perform, we design a knowledge-based prompt as shown in Fig \ref{model}. Each input sequence of all the tasks follows the same pattern\footnote{Here, we only display the translated English version.}:
\begin{equation}
    \underbrace{\textbf{[TASK]} \mathit{name}}_{task\ prompt} \ \underbrace{\textbf{[DIALOGUE]} \mathit{input\ utterances}}_{dialogue\ history} \ \underbrace{\textbf{[GOAL]}\mathit{description} }_{goal\ prompt}
\end{equation}
The task prompt denotes task names such as domain classification and dialogue generation to distinguish different pre-training tasks. And the goal prompt represents task instructions that describe the goal the task aims to accomplish. Combining task and goal prompt bridges connections between different tasks and avoids extreme dependency of the model on specific tasks to learn disentangled task representations. We find both prompts contribute to the final performance in Section \ref{prompt}. Besides, describing tasks via natural language helps our model has a strong generalization capability to unseen downstream tasks (see Section \ref{Generalization}). To add role information of input sequence, we append special tokens \emph{[CUSTOMER]} or \emph{[AGENT]} before each utterance. For the domain and intent tasks, we only select the first two utterances of a user as dialogue history and predict the label words. For the dialogue generation task, we split the whole conversation into multiple training segments and predict each agent's utterance. For the summarization task, we take the overall dialogue history as input and predict the summary. We use the same maximum likelihood objective to iteratively optimize these dialogue tasks.

\section{Experiment}

\subsection{Datasets}
\begin{table}[t]
\caption{The statistics of the experimental datasets.}
\label{dataset}
\resizebox{0.46\textwidth}{!}{
\begin{tabular}{l|cccc}
\hline
Datasets              & Train   & Dev   & Test   & Number of classes \\ \hline
Domain Classification & 139,209 & 1,000 & 19,885 & 24                \\
Intent Detection      & 18,342  & 1,000 & 3,725  & 20                \\
Dialogue Generation   & 39,315  & 1,000 & 2,087  & -                 \\
Summarization         & 44,681  & 1,000 & 3,727  & -                \\ \hline
\end{tabular}}
\end{table}

We evaluate the effectiveness of UFA in four downstream tasks: domain classification, intent detection, dialogue generation, and summarization. Note that these finetuning datasets have no overlap with the pre-training corpus and are 100\% labeled by human annotators. For classification tasks, we use accuracy as the evaluation metric. For generation tasks, we use bleu-2 and rouge scores as the evaluation metrics. Table \ref{dataset} shows the statistics of the datasets. For the training set, we hold out 1000 samples as the validation set to select the best model in fine-tuning. We remove all the punctuation in the label words and output the prediction strings. For classification tasks, we exactly match the outputs with the original labels and regard them as correct only if the predicted tokens are the same as the label words. In practice, we find UFA generates almost the same word sequences as label space.

\subsection{Baselines and Implementation Details}
We compare our method with the following baselines:
\begin{itemize}
    \item \textbf{Bert-base}: the widely used Chinese pre-trained language model proposed by Google \footnote{https://github.com/google-research/bert}. It has 12-layer, 768-hidden, 12-heads, 110M parameters.
    \item \textbf{Roberta-large}: Since Google doesn't release Bert-large version for Chinese, we adopt another Chinese Roberta-large from \citet{cui-etal-2020-revisiting} \footnote{https://github.com/ymcui/Chinese-BERT-wwm}. It has 24-layer, 1024-hidden, 16-heads, 330M parameters. Note that these two models have different pre-training corpus and training settings.
    \item \textbf{mT5}: A massively multilingual pre-trained transformer model using similar recipe as T5 \cite{Raffel2020ExploringTL}. We use its small version (8 layers, 6 heads, 512 hidden, 300M parameters).
    \item \textbf{UFA-ori}: a variant which is  pre-trained only on the unlabeled dialogue corpus.
\end{itemize}
In our experiments, we use a dense left-to-right, encoder-decoder transformer language model of 8-layer, 6-head, 512-hidden, 300M parameters similar to mT5. We use the SentencePiece library \cite{kudo-richardson-2018-sentencepiece} to tokenize dialogues with a 60,108 vocabulary which is smaller than mT5 (250,112). We keep only Chinese and English characters. We finetune all models for 20 epochs with a batch size of 32 using the Adafactor Optimizer \cite{Shazeer2018AdafactorAL} with a learning rate of 1e-4. The input and target sequence lengths used in finetuning are 512 and 100.

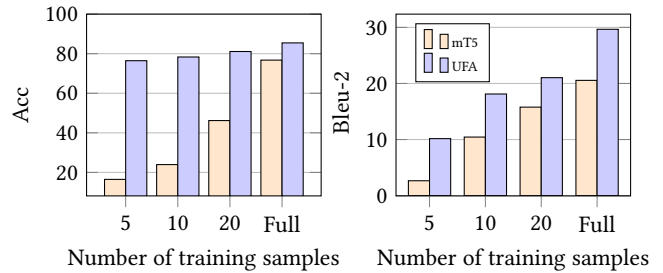
\begin{figure}[t]
    \raggedright
    \begin{minipage}[b]{0.4\linewidth}
    \centering
    \vspace{0pt}
	\begin{tikzpicture}
	\begin{axis}[ybar,enlarge x limits=0.25,xtick={5,10,20,Full},symbolic x coords={5,10,20,Full},ymajorgrids=true,height=4cm,width=4.7cm,xtick pos=left,ymax=100,xlabel=Number of training samples,ylabel=Acc,bar width=8pt]
	\addplot[draw=black,fill=orange!20,bar shift=-4pt]
	coordinates
	{(5,16.48)
    (10,23.95)
    (20,46.20)
    (Full,76.75)
};
	\addplot[draw=black,fill=blue!20,bar shift=4pt]
	coordinates
	{(5,76.45)
    (10,78.33)
    (20,81.10)
    (Full,85.45)};
	\end{axis}
	\end{tikzpicture}
\end{minipage}
\hspace{20pt}
\begin{minipage}[b]{0.4\linewidth}
\centering
\vspace{0pt}
	\begin{tikzpicture}
	\begin{axis}[ybar,enlarge x limits=0.2,xtick={5,10,20,Full},symbolic x coords={5,10,20,Full},ymajorgrids=true,height=4cm,width=4.7cm,xtick pos=left,legend style={at={(0.4,0.95)},nodes={scale=0.6}},xlabel=Number of training samples,ylabel=Bleu-2,bar width=8pt,ylabel style={xshift=1pt}]
	\addplot[draw=black,fill=orange!20,bar shift=-4pt]
	coordinates
	{(5,2.65)
    (10,10.44)
    (20,15.78)
    (Full,20.55)
};
	\addplot[draw=black,fill=blue!20,bar shift=4pt]
	coordinates
	{(5,10.16)
    (10,18.12)
    (20,21.03)
    (Full,29.66)};
	\legend{mT5,UFA}
	\end{axis}
	\end{tikzpicture}
\end{minipage}
\vspace{-0.45cm}
\caption{Effectiveness of UFA under the few-shot setting.}
\label{few}
\vspace{-0.25cm}
\end{figure}

\subsection{Main Results}

Table \ref{main} displays the main results of our UFA compared to baselines on four downstream dialogue tasks. Our method UFA significantly outperforms the baselines by 2.69\% accuracy on NLU tasks and 4.79\% Rouge-1 score on NLG tasks. The results demonstrate the effectiveness of our proposed unified knowledge prompt pre-training framework. Specifically, for NLU tasks, UFA outperforms the bert-base \footnote{Roberta-large and bert-base we use come from different open source projects, so the pre-training settings are not the same, including corpus and training steps, which may cause the performance difference.} by 1.93\% on domain classification and 3.44\% on intent detection. We find generative PTMs like mT5 and UFA-ori gets lower accuracy than bert or roberta, which indicates bert-based NLU models have an advantage over raw generative models. Performing knowledge prompt pre-training improves 4.32\% accuracy compared to UFA-ori. For NLG tasks, UFA outperforms mT5 by 9.11\% Bleu-2 and 6.97\% Rouge-1 on dialogue generation and 2.61\% Rouge-1 on summarization. We find knowledge pre-training obtains larger improvements on dialogue generation. We argue generation space of dialogue is much larger than summarization and further pre-training helps the model explore diverse action spaces. But summarization in customer service dialogues focuses on extracting limited key points from the dialogue history and won't benefit much from large-scale pre-training. Overall, under similar parameters, our proposed UFA gets consistent improvements than baselines on all tasks.

\section{Analysis}

\subsection{Few-Shot Learning}

\begin{table}[t]
\caption{Generalization analysis of UFA on the unseen sentence similarity task.}
\label{unseenr}
\vspace{-0.35cm}
\resizebox{0.38\textwidth}{!}{
\begin{tabular}{l|llll} 
\hline
              & \multicolumn{1}{c}{Acc}   & \multicolumn{1}{c}{F1}    & \multicolumn{1}{c}{P}     & \multicolumn{1}{c}{R}     \\ \hline
bert-base     & \multicolumn{1}{c}{81.80} & \multicolumn{1}{c}{81.80} & \multicolumn{1}{c}{81.80} & \multicolumn{1}{c}{81.80} \\
roberta-large & 80.68                     & 80.50                     & 80.91                     & 80.12                     \\
mT5           & 73.63                     & 73.34                     & 75.16                     & 73.87                     \\ \hline
UFA-ori       & 82.03                     & 82.00                     & 82.12                     & 81.98                     \\
UFA           & \textbf{87.50}                     & \textbf{87.49}                     & \textbf{87.52}                     & \textbf{87.48}                    \\ \hline
\end{tabular}}
\vspace{-0.5cm}
\end{table}

To verify the effectiveness of UFA under the few-shot setting, we perform an analysis in Fig \ref{few}. We select intent detection and dialogue generation tasks for space limitation and randomly choose 5, 10, and 20 samples for each class. Since dialogue generation doesn't have a fixed label set, we choose  5, 10 and 20 (context, response) training samples instead. We use the original dev set and test set for evaluation. The results show that UFA obtains larger improvements in the few-shot learning. For intent detection, UFA achieves comparable performance to full training data using 20 examples. For dialogue generation, both methods require adequate data to get superior performance, which indicates the higher complexity of dialogue generation. Pre-training on the weak-supervised data helps the model learn domain-adaptive expert knowledge and reduce the labeling cost of fine-tuning downstream tasks. Exploring zero-shot learning without any target label is valuable for future work.

\subsection{Generalization to Unseen Task}
\label{Generalization}
We perform a generalization analysis of UFA on the unseen task in Table \ref{unseenr} to figure out how UFA performs on tasks not seen in knowledge prompt pre-training. We establish a sentence similarity dataset that identifies whether two input user queries are relevant or not. The number of training, dev and test set is 10000, 1000 and 3000. We report accuracy and macro Precision/Recall/F1 metrics. For UFA and UFA-ori, we construct the same prompt as Fig \ref{model} where the task name is \emph{sentence similarity} and the goal description is \emph{the relationship of the input sentences is}. The output labels are \emph{positive} or \emph{negative}. We find UFA outperforms the baselines with a large margin of 5.70\% Acc and 5.69\% F1, which confirms UFA learns generalized dialogue representations by knowledge prompt pre-training and has strong transferability to out-of-distribution (OOD) tasks. Generalization capability is vital to practical application in customer service dialogues where new questions and scenarios are evolving. Compared to UFA-ori, UFA gets significant improvements. It proves traditional unsupervised pre-training on unlabeled dialogue corpus suffers from good adaptation to downstream tasks since expert knowledge is quite important.

\subsection{Effect of Prompt}
\label{prompt}
Table \ref{promptr} shows the effect of different prompts in UFA finetuning. We separately remove each part of the UFA prompt and see how it affects. Results show that removing the task or goal prompt substantially decreases the performance of the dialogue generation task but has a subtle effect on the intent detection task. We assume high complexity of generation tasks requires more handcrafted prompt templates. Comparing the two prompts, the goal prompt gains superior metrics on the generation task but is similar to the task prompt on the classification task. Natural language instructions like goal prompt can help the model share knowledge across tasks, especially for hard generation tasks.

\begin{table}[t]
\caption{Effect of different prompts.}
\label{promptr}
\vspace{-0.35cm}
\resizebox{0.42\textwidth}{!}{
\begin{tabular}{l|c|cc}
\hline
           & \multicolumn{1}{l|}{Intent Detection} & \multicolumn{2}{l}{Dialogue Generation} \\
                            & Acc                                  & Bleu-2             & Rouge-1            \\ \hline
UFA                         & 85.45                                & 29.66              & 48.30              \\
-w/o goal prompt            & 85.23                                & 28.11              & 44.94              \\
-w/o task prompt            & 85.12                                & 28.40              & 45.92              \\ \hline

\end{tabular}}
\vspace{-0.5cm}
\end{table}

\begin{table}[t]
\caption{Effect of different pre-training tasks.}
\label{ablation}
\resizebox{0.42\textwidth}{!}{
\begin{tabular}{l|ccc}
\hline
Pre-training tasks & \multicolumn{1}{l}{Intent Detection} & \multicolumn{2}{l}{Dialogue Generation} \\
                  & Acc                                  & bleu-2             & rouge-1            \\ \hline
UFA-ori           & 79.44                                & 27.20              & 45.93              \\
+domain           & 80.18                                & 29.15              & 48.15              \\
+intent           & 85.12                                & 25.43              & 39.51              \\
+summary          & 84.85                                & 29.09              & 40.54              \\
+dialogue         &      83.91 
&
28.70 &

46.23                                                  \\ \hline
UFA               & \textbf{85.45}                       & \textbf{29.66}     & \textbf{48.30}   \\ \hline 
\end{tabular}}
\vspace{-0.45cm}
\end{table}

\subsection{Ablation Study}
Table \ref{ablation} shows the results for each pre-training task. We find each task contributes to the part of final results and our proposed UFA gets the best overall performance. It proves a unified prompt framework helps disentangle task dependency and bridge connections between different tasks to avoid imbalance issues.

\section{Conclusion}
In this paper, we propose a novel unified knowledge prompt pre-training framework, UFA, for customer service dialogues. Different from existing work, UFA aims to learn expert knowledge from weakly-annotated dialogue labels instead of linguistic knowledge from unsupervised dialogue corpus. We formulate all the tasks of customer service dialogues as a unified text-to-text generation task and introduce a knowledge-driven prompt strategy to jointly learn from multiple dialogue tasks. We pre-train UFA on a large-scale Chinese customer service corpus collected from practical scenarios. The results on four downstream tasks show the effectiveness of UFA. We also find UFA has strong few-shot learning and generalization capability. 





\bibliographystyle{ACM-Reference-Format}
\balance
\bibliography{cikm}

\end{document}